\documentclass{article}

\usepackage{arxiv}
\usepackage{amsmath, amssymb}
\usepackage{algorithm}
\usepackage{algorithmic}
\usepackage{authblk}
\usepackage{multirow}
\usepackage{tabularx}
\usepackage{pifont}
\usepackage{makecell}
\usepackage[table]{xcolor}
\usepackage[utf8]{inputenc} 
\usepackage[T1]{fontenc}    
\usepackage{hyperref}       
\usepackage{url}            
\usepackage{booktabs}       
\usepackage{amsfonts}       
\usepackage{nicefrac}       
\usepackage{microtype}      
\usepackage{lipsum}
\usepackage{graphicx}
\graphicspath{ {./images/} }
\newcommand{\cmark}{\ding{51}}  
\newcommand{\xmark}{\ding{55}}  

\title{ST-Prune: Training-Free Spatio-Temporal Token Pruning for Vision-Language Models in Autonomous Driving}

\author[1]{Lin Sha}
\author[1,2*]{Haiyun Guo}
\author[3]{Tao Wang}
\author[3]{Cong Zhang}
\author[1]{Min Huang}
\author[1,2]{Jinqiao Wang}
\author[1*]{Qinghai Miao}

\affil[1]{School of Artificial Intelligence, University of Chinese Academy of Sciences, Beijing 100049, China}
\affil[2]{Institute of Automation, Chinese Academy of Sciences, Beijing 100190, China}
\affil[3]{Carizon, Beijing 100094, China}

\begin{document}
\maketitle
\begin{abstract}
Vision-Language Models (VLMs) have become central to autonomous driving systems, yet their deployment is severely bottlenecked by the massive computational overhead of multi-view camera and multi-frame video input. Existing token pruning methods, primarily designed for single-image inputs, treat each frame or view in isolation and thus fail to exploit the inherent spatio-temporal redundancies in driving scenarios. To bridge this gap, we propose ST-Prune, a training-free, plug-and-play framework comprising two complementary modules: Motion-aware Temporal Pruning (MTP) and Ring-view Spatial Pruning (RSP). MTP addresses temporal redundancy by encoding motion volatility and temporal recency as soft constraints within the diversity selection objective, prioritizing dynamic trajectories and current-frame content over static historical background. RSP further resolves spatial redundancy by exploiting the ring-view camera geometry to penalize bilateral cross-view similarity, eliminating duplicate projections and residual background that temporal pruning alone cannot suppress. These two modules together constitute a complete spatio-temporal pruning process, preserving key scene information under strict compression. Validated across four benchmarks spanning perception, prediction, and planning, ST-Prune establishes new state-of-the-art for training-free token pruning. Notably, even at 90\% token reduction, ST-Prune achieves near-lossless performance with certain metrics surpassing the full-model baseline, while maintaining inference speeds comparable to existing pruning approaches.
\end{abstract}


\section{Introduction}

The realization of safe and reliable autonomous driving(AD) necessitates a system capable of simultaneous high-fidelity perception, complex behavioral reasoning, and real-time control within highly dynamic environments ~\cite{chen2019autonomous,hu2026advanced, yang2025trajectory}. This multi-faceted cognitive demand has driven the rapid adoption of Vision-Language Models (VLMs) as the central reasoning backbone of modern AD systems~\cite{zhou2024vision,tian2025large}. By leveraging the extensive world knowledge encapsulated within Large Language Models (LLMs), these frameworks facilitate sophisticated scene understanding~\cite{hu2022enhancing,predicting,muhammad2022vision} and closed-loop decision-making~\cite{zhang2022rethinking,jia2024bench2drive,ljungbergh2024neuroncap} within a unified architecture.

Despite this rapid progress, this powerful paradigm faces a formidable \textit{Curse of Dimensionality}, a bottleneck highlighted by Tesla’s AI leadership\cite{elluswamy2025tesla}. Specifically, this challenge is characterized by a two-fold compounding dimension. First, unlike general VLM applications that typically process isolated images, an autonomous vehicle must integrate synchronized inputs from a multi-view camera suite (e.g., six or more views) across a continuous temporal horizon. This results in the generation of thousands of visual tokens per inference step, creating an extreme many-to-few mapping where massive high-dimensional input must be compressed into a sparse set of trajectory waypoints or steering commands. Second, there exists a profound imbalance in the semantic density of these tokens. In standard driving sequences, the vast majority of visual data represents redundant background scenery or static road geometry, contributing negligibly to the immediate driving policy. Conversely, safety-critical long-tail events such as a pedestrian obscured by a vehicle or a transient traffic signal change—occupy an infinitesimally small fraction of the total token space\cite{corner}, yet represent the most vital cues for navigation. Together, these two dimensions force the model to expend significant computational resources on low-entropy tokens while struggling to explore the rare, salient information required for safety driving decision.

\begin{figure}[!t]
  \centering
  \includegraphics[width=\linewidth]{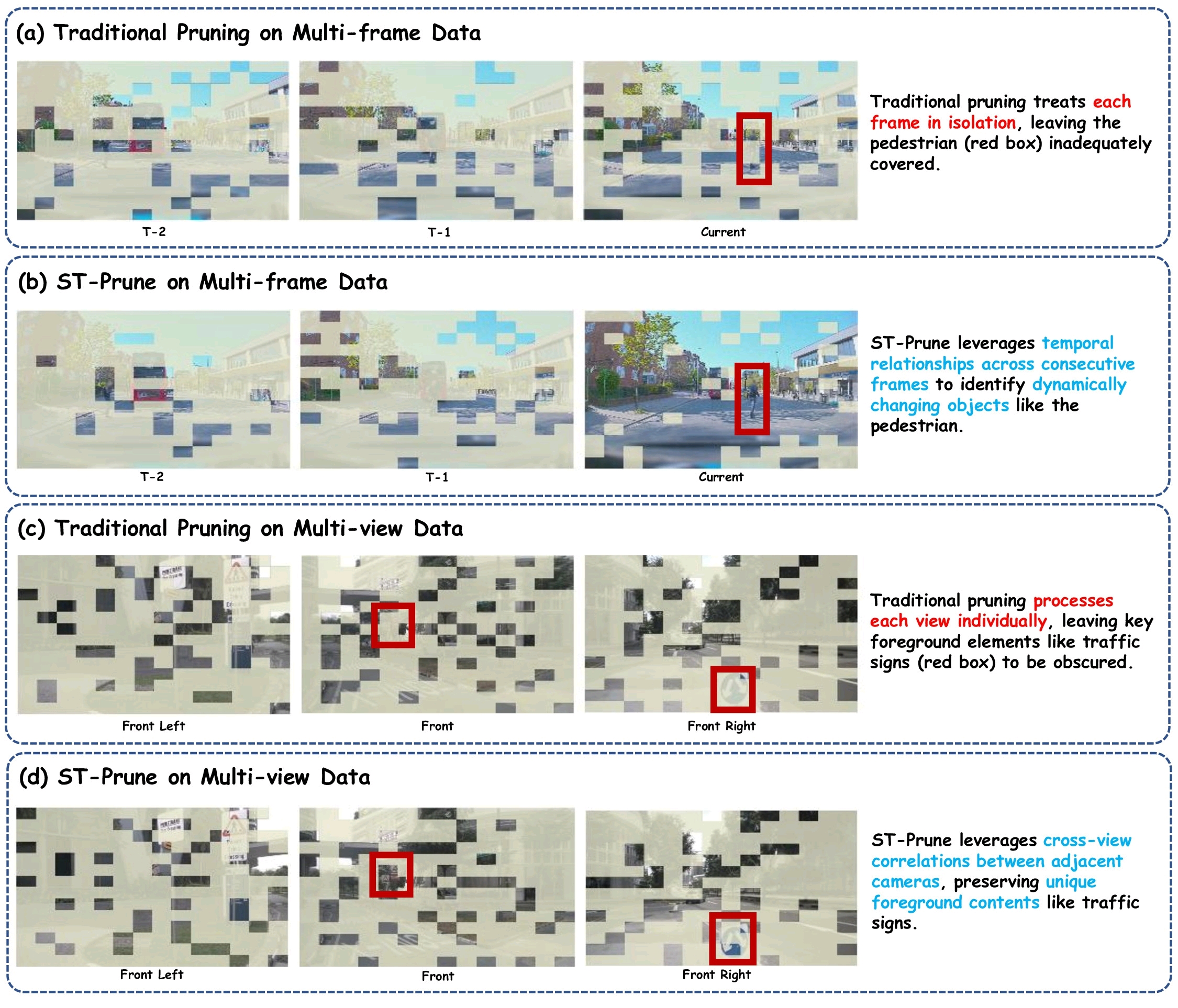}
\caption{Illustration of the failure modes of traditional pruning 
methods on spatio-temporal driving inputs and the corresponding 
improvements achieved by ST-Prune.}
  \label{fig:motivation}
\end{figure}

To address this computational overhead, visual token pruning has emerged as a pivotal research frontier. Pioneering works such as the attention-based FAST-V~\cite{fastv} and the similarity-based ToMe~\cite{tome} have demonstrated that redundant tokens can be substantially eliminated with minimal impact on model performance, inspiring a growing body of subsequent methods~\cite{sparsevlm,wen2025stop, wen2025token} that further advance token reduction across general-purpose VLMs. However, these methods are designed exclusively for single-image inputs and operate on a flat token space with no awareness of temporal ordering or omnidirectional camera configuration, rendering them fundamentally ill-suited for AD. As illustrated in Figure~\ref{fig:motivation}, this leads to two concrete failure modes. Temporally, treating each frame in isolation without exploiting inter-frame temporal relationships distributes the token budget uniformly across all frames, allowing static historical background to crowd out current dynamic objects such as pedestrians. Spatially, treating camera views in isolation allows cross-view overlapping background regions to monopolize the token budget, marginalizing unique foreground content such as traffic signs that occupy only a small fraction of the overall visual field. Within the AD domain, FastDriveVLA~\cite{fastdrivevla} and Prune2Drive~\cite{prune2drive} represent the only two pruning frameworks specifically tailored to driving scenarios, employing MAE-based reconstruction~\cite{mae} and Bayesian parameter search respectively to address redundancy across camera views. However, both methods introduce significant overhead through additional training phases or expensive offline search processes, and neither provides a unified solution simultaneously resolving redundancy across both temporal and spatial dimensions.

To bridge this gap, we propose ST-Prune, a truly training-free, plug-and-play framework comprising two complementary modules that jointly address spatio-temporal redundancy in autonomous driving. Motion-aware Temporal Pruning (MTP) leverages motion volatility and temporal recency to prioritize dynamic trajectories and recent observations, while aggressively discarding static redundancy from historical frames. Ring-view Spatial Pruning (RSP) exploits the physical spatial structure of surround-view cameras to penalize bilateral cross-view similarity, eliminating redundant background tokens that appear near-identically across adjacent views and thereby redirecting the pruning budget toward unique foreground content. As shown in Figure~\ref{fig:motivation}, ST-Prune correctly retains the pedestrian in the current frame through temporal recency bias and achieves more complete coverage of the traffic sign across adjacent views by suppressing cross-view redundant background. Furthermore, ST-Prune requires no training, no calibration data, and no architectural modifications, and is fully compatible with efficient kernels such as FlashAttention and KV cache optimization. The main contributions of this paper are summarized as follows:
\begin{itemize}

\item We present ST-Prune, the first training-free, plug-and-play framework that simultaneously addresses token redundancy across both the multi-frame temporal dimension and the multi-view spatial dimension in autonomous driving. 

\item We propose MTP and RSP, two complementary modules oriented for temporal and spatial redundancy in AD
respectively. We further demonstrate that each module serves as a robust standalone solution, delivering consistent improvements across diverse driving scenarios.

\item  We validate ST-Prune across four mainstream AD benchmarks, namely DriveLM, LingoQA, NuInstruct, and OmniDrive, consistently outperforming existing pruning methods across perception, prediction, and planning tasks under both 25\% and 10\% token retention rates, while maintaining comparable inference efficiency. 
\end{itemize}
\section{Related work}

\subsection{Vision-Language Driving Models}
The integration of LLMs into AD has progressed rapidly through several distinct phases~\cite{chen2023feedback}. Early pioneering works such as GPTDriver~\cite{gptdriver} and Driving with LLMs~\cite{driving-with-llms} demonstrated the feasibility of leveraging LLMs for driving decisions, establishing the foundation for subsequent VLM-based approaches. Building upon, a series of works explored using VLMs as supervisory signals or auxiliary modules within end-to-end driving pipelines\cite{LMDrive, senna, vlm-ad}. With the emergence of Vision-Language Action(VLA) models, VLMs have since evolved into the central cognitive unit of autonomous vehicles, directly generating navigation decisions and control signals within a unified framework\cite{orion, covla, drivemoe}.
Parallel to this progression, researchers have identified key limitations of general VLMs when applied to driving scenarios and proposed targeted enhancements like spatial grounding\cite{vggdrive, mpdrive, spacedrive} and logical reasoning\cite{reason2drive, dynvla, fastdrivecot}. Another distinct branch focuses on unified multi-task architectures: rather than addressing individual limitations, DriveMM~\cite{drivemm} proposes an all-in-one large multimodal model that processes diverse driving inputs and achieves state-of-the-art performance across a broad array of benchmarks through sophisticated cross-modal alignment strategies.

Overall, these studies demonstrate the strong potential of VLMs as a unified paradigm for perception, reasoning, and decision-making in autonomous driving. However, the growing capability of VLM-based driving systems comes at a steep computational cost, especially when processing dense spatio-temporal visual tokens in real-world scenes.

\subsection{Token Pruning for Vision-Language Models}
To mitigate the quadratic computational complexity of Transformer-based VLMs, visual token pruning has emerged as a pivotal research direction, broadly categorized into attention-based, similarity-based, and hybrid approaches. FAST-V~\cite{fastv} leverages decayed attention scores in deeper layers as saliency proxies to filter low-contribution tokens. A more principled direction is represented by similarity-based methods: DivPrune~\cite{divprune} reformulates token reduction as a max-min diversity problem to select a complementary, non-redundant token subset, while VisPruner~\cite{vispruner} combines visual self-attention with similarity-based de-duplication to overcome the limitations of purely cross-modal attention signals. Recognizing that pruned tokens may carry recoverable information, hybrid approaches such as SparseVLM~\cite{sparsevlm} and PACT~\cite{pact} further integrate clustering mechanisms to recycle rather than simply discard redundant tokens. Despite their effectiveness on general VLM benchmarks, all these methods are designed for single-image inputs and remain fundamentally agnostic to the spatio-temporal structure of autonomous driving.

Although FastDriveVLA~\cite{fastdrivevla} and Prune2Drive~\cite{prune2drive} have made pioneering attempts to tailor token pruning for driving scenarios through MAE-based reconstruction and Bayesian parameter search respectively, their applicability remains fundamentally constrained. 
On one hand, the additional retraining and offline search processes they require introduce non-trivial deployment barriers that undermine their plug-and-play utility. On the other hand, both approaches confine their pruning objectives to spatial redundancy within individual timestamps, leaving the temporal dimension entirely unaddressed. 
This reveals a critical and persistent void in the field: a truly training-free framework capable of jointly suppressing redundancy across both the multi-frame temporal axis and the multi-view spatial axis has yet to emerge.

\section{Problem Formulation}\label{sec:problem}

Modern VLM-based autonomous driving systems are equipped with $V$ synchronized surround-view cameras, each capturing $T$ consecutive frames. Each frame is processed by a vision encoder followed by a projector, producing $P$ spatial patch tokens of embedding dimension $C$ per view per frame. Without any pruning, the total token count fed into the LLM backbone amounts to $V \times T \times P$, which grows rapidly with the number of cameras and frames, imposing prohibitive computational overhead.

In practice, the prompt design of VLM-based driving systems explicitly encodes the camera identity of each token, informing the model which tokens belong to which view. This means each camera view must be preserved as a distinct perceptual unit — discarding entire views would break the correspondence between visual tokens and their camera identities, resulting in critical blind spots and corrupting the spatial awareness of the model. This naturally imposes a per-view pruning constraint: rather than selecting tokens globally across all $V \times T \times P$ tokens, the objective is to retain $K$ tokens independently within each view, reducing the total token count from $V \times T \times P$ to a compact $V \times K$ with $K \ll T \times P$.

A direct and feasible approach to this problem is diversity maximization: within each view, select $K$ tokens from $T \times P$ such that the selected subset is maximally diverse in feature space, thereby suppressing redundancy while retaining representative content. Formally, this requires solving:
\begin{equation}
    \mathcal{S}^* = \arg\max_{|\mathcal{S}|=K} 
    \min_{\mathbf{x}_i, \mathbf{x}_j \in \mathcal{S}, i \neq j} 
    d(\mathbf{x}_i, \mathbf{x}_j),
\end{equation}
where $d(\cdot, \cdot)$ denotes a distance metric. However, this combinatorial optimization is NP-hard in general. To make it tractable, DivPrune~\cite{divprune} proposes a greedy approximation that iteratively selects the token maximizing its minimum distance to all previously selected tokens:
\begin{equation}
    S_{D} = \arg\max_{\mathbf{x}_i \notin \mathcal{S}} 
    \min_{\mathbf{x}_j \in \mathcal{S}} d(\mathbf{x}_i, 
    \mathbf{x}_j).
\end{equation}
This greedy formulation is training-free, plug-and-play, and produces a well-distributed token subset that naturally suppresses redundancy within a flat token space.

Although this approach appears feasible, it remains agnostic to the structural semantics inherent in the two key dimensions of the driving input. Along the temporal dimension $T$, tokens from the current frame are strictly more actionable than those from distant historical frames, yet vanilla max-min selection assigns them equal priority, allowing stale background content to compete with critical current-frame observations. Along the spatial dimension $V$, tokens appearing near-identically across adjacent camera views represent physical redundancy arising from overlapping fields of view, which flat per-view diversity selection operating independently within each view has no mechanism to detect or exploit. These two structural blindspots motivate ST-Prune, which reformulates this greedy selection as a domain-constrained diversity maximization problem, incorporating temporal and geometric priors of autonomous driving directly into the selection objective.

\begin{figure*}[!t]
  \centering
  \includegraphics[width=\linewidth]{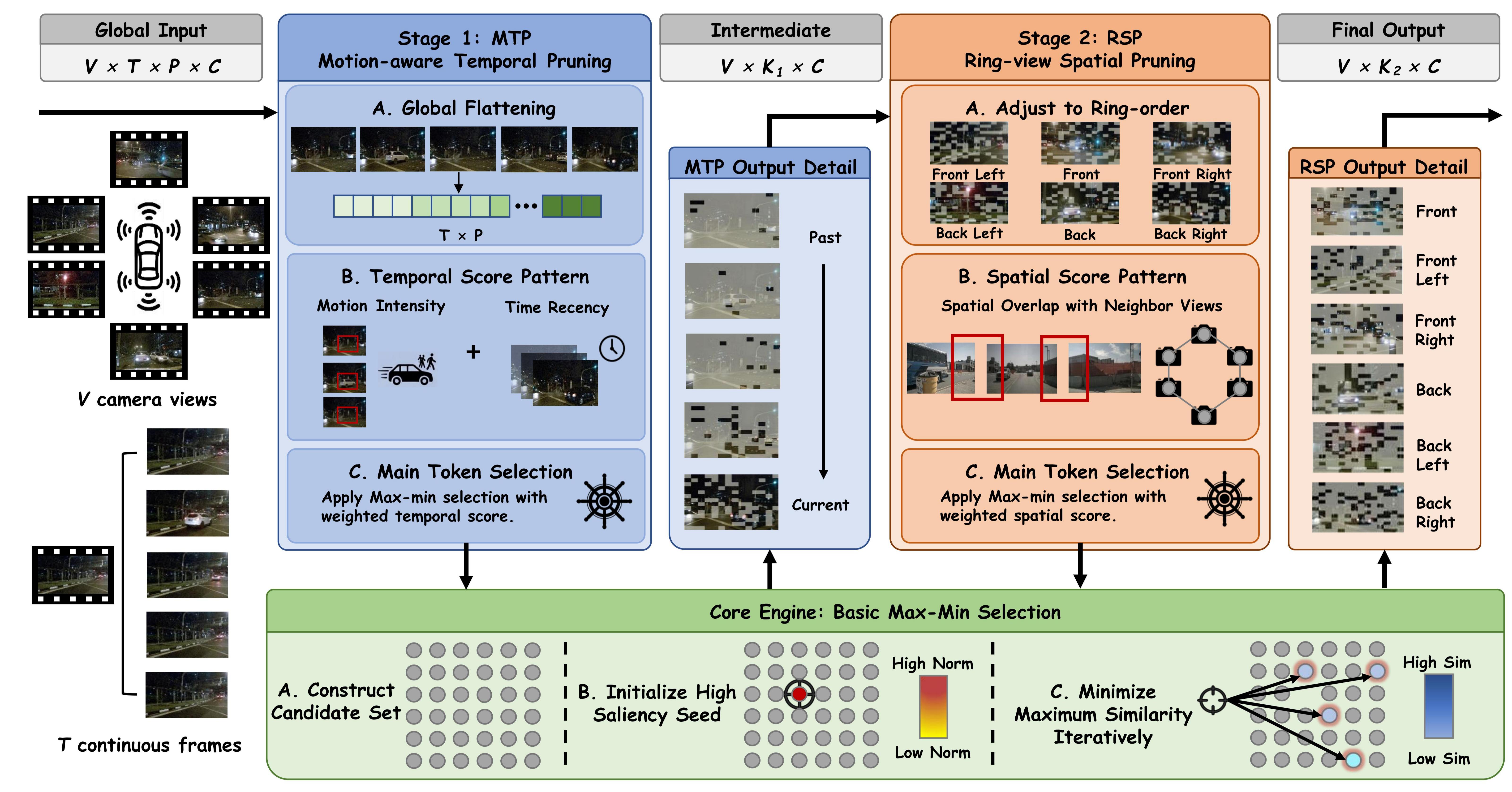}
  \caption{The ST-Prune pipeline for spatio-temporal token reduction. The framework processes multi-view, multi-frame inputs through a dual-stage selection process: (1) MTP, which targets temporal redundancy using motion and recency priors, and (2) RSP, which addresses spatial redundancy across the multi-view spatial geometry. Each stage leverages an iterative Max-Min Selection engine to maintain a diverse and semantically rich token set by optimizing a diversity objective augmented with corresponding temporal or spatial weighted scores.}
  \label{fig:overview}
\end{figure*}

\section{Methodology}

\subsection{Method Overview}

ST-Prune serves as a model-agnostic, plug-and-play framework integrated within the latent processing phase of the LLM. It requires no structural modifications to the backbone or additional training. As illustrated in Figure~\ref{fig:overview}, given a high-dimensional spatio-temporal input tensor $\mathbf{X} \in \mathbb{R}^{V \times T \times P \times C}$, the framework operates through two sequential stages built upon a shared Core Max-Min Selection Engine.

In the first stage, Motion-aware Temporal Pruning(MTP) addresses temporal redundancy across $T$ frames. For each camera view, the temporal token sequence is globally flattened into $T \times P$ tokens, upon which a temporal score pattern combining motion intensity and temporal recency is computed. The weighted max-min selection then retains $K_1$ tokens per view, producing an intermediate representation $\mathbf{X}_1 \in \mathbb{R}^{V \times K_1 \times C}$ that preserves dynamic trajectories and critical current-frame content while aggressively discarding static historical background.

In the second stage, Ring-view Spatial Pruning(RSP) addresses spatial redundancy across $V$ camera views. The intermediate tokens are first rearranged into ring order — \texttt{Front\_Left}, \texttt{Front}, \texttt{Front\_Right}, \texttt{Back\_Right}, \texttt{Back}, \texttt{Back\_Left} — to reflect the physical adjacency of the surround-view camera array. A bilateral spatial score pattern is then computed for each token by measuring its similarity to tokens in both neighboring views, identifying cross-view redundancy arising from overlapping fields of view. The weighted max-min selection subsequently operates independently within each view, retaining $K_2$ tokens per view to produce the final output $\mathbf{X}_2 \in \mathbb{R}^{V \times K_2 \times C}$, reducing the total token count from $V \times T \times P$ to $V \times K_2$.

Together, the two stages form a coarse-to-fine pipeline that collectively suppresses spatio-temporal redundancy: MTP performs a coarse temporal distillation across frames, while RSP performs a fine-grained spatial refinement across views, each guided by domain-specific structural priors rather than generic feature statistics. In the following subsections, we provide a formal mathematical exposition of MTP and RSP, detailing their respective scoring functions and their integration into the core selection engine.

\subsection{Motion-aware Temporal Pruning}

We represent the visual tokens of a video sequence as a temporal tensor $\mathbf{X}_{\text{tem}} \in \mathbb{R}^{T \times P \times C}$. While aggressive pruning of static background tokens is desirable, a naive focus on motion intensity alone may inadvertently discard essential stationary landmarks such as traffic lights and stop signs in the current frame, which are indispensable for immediate decision-making. To address this, we propose Motion-aware Temporal Pruning (MTP), which reformulates the unconstrained diversity objective into a temporally-aware selection problem by introducing a unified scoring function that encodes two domain-specific priors as soft constraints: motion volatility, derived from the temporal variance of patch features, and temporal recency, modeled as an exponential bias toward the current frame. 

Formally, for each token $\mathbf{x}_{t,p}$ at frame $t$ and spatial patch $p$, the temporal importance score is defined as:
\begin{equation}
S_{\text{tem}}(t,p) = \|\mathbf{x}_{t,p} - \bar{\mathbf{x}}_p\|_2^2 + \exp\left(\alpha \cdot \frac{t}{T}\right)
\end{equation}
where $\bar{\mathbf{x}}_p$ denotes the temporal mean of patch features at position $p$, and $\alpha$ is a scaling factor that controls the exponential emphasis on the most recent frame—set to 2 throughout our experiments. Both terms are independently normalized via min-max normalization before summation to prevent either component from dominating due to scale differences. Prior to selection, the spatio-temporal tensor is flattened into a sequence of $N = T \times P$ tokens to enable a global search across the entire video volume.

The selection of $\mathcal{S}$ proceeds in a seed-then-expand strategy. Vanilla max-min selection initializes the seed by selecting the token with maximum similarity to all others, anchoring to the most representative token in feature space. However, this initialization is agnostic to token informativeness and temporal relevance. Instead, we introduce feature norm as a proxy for token importance to identify the most informative anchor, and further incorporate the temporal scoring function to bias the selection toward dynamic and recent-frame content:
\begin{equation}
s_1 = \arg\max_{\mathbf{x}_i \in \mathbf{X}_{\text{tem}}} 
\left[\|\mathbf{x}_i\|_2 + \lambda_1 \cdot S_{\text{tem}}(\mathbf{x}_i)
\right]
\end{equation}
where $\lambda_1$ is a balancing coefficient. This ensures that the initial anchor is not merely a central feature vector but one that is simultaneously salient and temporally relevant, orienting the entire subsequent selection sequence toward dynamic and recent-frame content from the very first step.

Subsequent tokens are selected by extending the max-min diversity criterion with a temporal awareness penalty: tokens that are both dissimilar to the existing set and temporally important are jointly favored:
\begin{equation}
s_m = \arg\min_{\mathbf{x}_i \notin \mathcal{S}} \left\{\max_{\mathbf{x}_j \in \mathcal{S}} \left[\text{sim}(\mathbf{x}_i, \mathbf{x}_j)\right] + \lambda_1 \cdot (1 - S_{\text{tem}}(\mathbf{x}_i))\right\}
\end{equation}
where cosine similarity is adopted as:
\begin{equation}
\text{sim}(\mathbf{x}_i, \mathbf{x}_j) = \frac{\mathbf{x}_i \cdot \mathbf{x}_j}{\|\mathbf{x}_i\| \|\mathbf{x}_j\|}
\end{equation}

Since the temporal recency term retains all content in the most recent frame indiscriminately — including residual static background — complementary spatial pruning is subsequently applied to complete the background suppression that temporal pruning alone cannot achieve.

\begin{algorithm}[t]
\footnotesize
\caption{Motion-aware Temporal Pruning (MTP)}
\label{alg:mtp}
\begin{algorithmic}[1]
\REQUIRE Temporal tensor $\mathbf{X}_{\text{tem}} \in 
\mathbb{R}^{T \times P \times C}$, temporal subset size $K_1$, balancing coefficient $\lambda_1$ 
\ENSURE Pruned tensor $\mathbf{X}_1 \in \mathbb{R}^{K_1 \times C}$
\STATE Flatten $\mathbf{X}_{\text{tem}}$ into $N = T \times P$ 
tokens
\STATE Compute $S_{\text{tem}}(t,p)$ for all tokens via Eq.~(3)
\STATE Apply min-max normalization to both terms of 
$S_{\text{tem}}$ independently
\STATE Initialize $\mathcal{S} \leftarrow \{s_1\}$ where $s_1$ 
is selected via Eq.~(4)
\FOR{$m = 2$ to $K_1$}
    \STATE Select $s_m$ via Eq.~(5); update $\mathcal{S} 
    \leftarrow \mathcal{S} \cup \{s_m\}$
\ENDFOR
\RETURN $\mathbf{X}_1 = \{\mathbf{x}_i \mid i \in \mathcal{S}\} 
\in \mathbb{R}^{K_1 \times C}$
\end{algorithmic}
\end{algorithm}

\subsection{Ring-view Spatial Pruning}
While MTP addresses temporal redundancy, autonomous driving systems exhibit significant spatial redundancy across the multi-camera suite, where adjacent cameras with overlapping fields of view simultaneously capture the same physical objects and background elements across neighboring views. We denote the visual tokens of such a surrounding-view system as a spatial tensor $\mathbf{X}_{\text{spa}} \in \mathbb{R}^{V \times P \times C}$, where $V$ represents the number of camera views.  Ring-view Spatial Pruning(RSP) incorporates the ring 
spatial arrangement of the annular camera layout as a 
geometric constraint on the diversity objective. Unlike naive similarity-based pruning that operates within a single view, RSP defines a bilateral spatial priority score that simultaneously queries both neighboring views in the ring, encoding the physical overlap structure of the camera geometry directly into the selection criterion. 

Specifically, to quantify cross-view redundancy, we define a spatial importance score for each token $\mathbf{x}_{v,p}$ in camera view $v$ and spatial patch $p$ as:
\begin{equation}
S_{\text{spa}}(v,p) = 1-\frac{1}{2}\left(\max_{k_1 \in P} 
\text{sim}(\mathbf{x}_{v,p}, \mathbf{x}_{v+1,k_1}) + \max_{k_2 \in P} 
\text{sim}(\mathbf{x}_{v,p}, \mathbf{x}_{v-1,k_2})\right)
\end{equation}
where indices are computed modulo $V$ to respect the ring topology, i.e., $\mathbf{x}_{V+1} \equiv \mathbf{x}_1$. A low $S_{\text{spa}}$ score indicates that a token is well-represented in both neighboring views and is therefore a strong candidate for removal. Unlike MTP which operates globally across all frames, RSP performs independent per-view selection to avoid erroneous cross-view comparisons between visually similar but spatially distinct objects, such as two black vehicles appearing in non-adjacent views.

The per-view selection follows the same seed-then-expand strategy as MTP, specialized with the spatial scoring function. The spatial seed for each view $v$ is identified as:
\begin{equation}
s_1(v) = \arg\max_{\mathbf{x}_i \in \mathbf{X}_{\text{spa},v}} 
\left[\|\mathbf{x}_i\|_2 + \lambda_2 \cdot 
S_{\text{spa}}(v, \mathbf{x}_i)\right]
\end{equation}
Subsequent tokens are selected by jointly maximizing diversity within 
the view and penalizing cross-view overlap:
\begin{equation}
s_m(v) = \arg\min_{\mathbf{x}_i \notin \mathcal{S}_v} \left\{
\max_{\mathbf{x}_j \in \mathcal{S}_v} \left[\text{sim}(\mathbf{x}_i, 
\mathbf{x}_j)\right] + \lambda_2 \cdot 
(1-S_{\text{spa}}(v, \mathbf{x}_i))\right\}
\end{equation}
By penalizing tokens with high bilateral cross-view similarity, RSP naturally identifies and removes not only duplicate object projections but also static background elements that appear near-identically across adjacent camera boundaries, completing the two-stage background suppression of ST-Prune.

\begin{algorithm}[t]
\footnotesize
\caption{Ring-view Spatial Pruning (RSP)}
\label{alg:rsp}
\begin{algorithmic}[1]
\REQUIRE Spatial tensor $\mathbf{X}_{\text{spa}} \in 
\mathbb{R}^{V \times P \times C}$, spatial subset size $K_2$, 
balancing coefficient $\lambda_2$
\ENSURE Pruned tensor $\mathbf{X}_2 \in \mathbb{R}^{V \times K_2 
\times C}$
\STATE Compute bilateral spatial score $S_{\text{spa}}(v,p)$ for 
all views and tokens via Eq.~(7)
\FOR{each camera view $v \in \{1, \dots, V\}$}
    \STATE Initialize $\mathcal{S}_v \leftarrow \{s_1(v)\}$ 
    where $s_1(v)$ is selected via Eq.~(8)
    \FOR{$m = 2$ to $K_2$}
        \STATE Select $s_m(v)$ via Eq.~(9); update 
        $\mathcal{S}_v \leftarrow \mathcal{S}_v \cup \{s_m(v)\}$
    \ENDFOR
    \STATE Collect selected tokens into $\mathbf{X}_{2,v}$
\ENDFOR
\RETURN $\mathbf{X}_2 = \{\mathbf{X}_{2,v}\}_{v=1}^{V} \in 
\mathbb{R}^{V \times K_2 \times C}$
\end{algorithmic}
\end{algorithm}

\subsection{Method Analysis}

\noindent\textbf{Complexity Analysis.}
The vanilla max-min selection incurs $O(K \cdot T \times P)$ 
complexity per view for selecting $K$ tokens from $T \times P$ 
candidates, yielding $O(V \cdot K \cdot T \cdot P)$ in total 
across all views. ST-Prune introduces two additional cost 
terms beyond this baseline. First, MTP performs iterative 
selection over $T \times P$ candidates for $K_1$ steps per 
view, incurring $O(V \cdot K_1 \cdot T \cdot P)$ --- 
identical in structure to vanilla selection and therefore 
introducing no asymptotic overhead relative to the baseline. 
Second, RSP introduces a one-time bilateral similarity 
precomputation of $O(V \cdot P^2)$ across neighboring views, 
after which per-view selection over the $K_1$ intermediate 
tokens costs $O(V \cdot K_2 \cdot K_1)$. The total complexity 
of ST-Prune is therefore:
\begin{equation}
O\left(V \cdot K_1 \cdot T \cdot P + V \cdot P^2 + 
V \cdot K_2 \cdot K_1\right)
\end{equation}
Since $K_2 \ll K_1 \ll T \cdot P$, the term $V \cdot K_2 \cdot K_1$ is negligible. The sole meaningful overhead relative to vanilla selection is the RSP precomputation $O(V \cdot P^2)$, which is computed once per inference pass and reused across all $K_2$ selection iterations. In practice, with $V \leqslant 7$ and $P$ bounded by the patch grid in vision encoder, this term remains fully manageable and comparable in magnitude to the vanilla selection cost, introducing no prohibitive overhead. This is empirically confirmed by the end-to-end latency measurements reported in Table~\ref{tab:efficiency_relative}.

\noindent\textbf{Module Ordering.}
A natural question is whether applying RSP before MTP could yield comparable results. While a spatial-first ordering is theoretically valid, it introduces a $T$-fold increase in RSP precomputation cost: bilateral similarity scores must be computed independently across all $T$ frames before MTP can reduce the temporal dimension, incurring $O(V \cdot T \cdot P^2)$ compared to $O(V \cdot P^2)$ in the temporal-first design. Furthermore, MTP naturally accommodates datasets without a ring-view camera structure by adjusting token ordering prior to RSP, whereas a spatial-first design would require an additional reordering step. These efficiency considerations motivate our temporal-first ordering. We empirically compare both orders in our ablation study(reported in Table~\ref{tab:ablation3}), which manifests that temporal-first consistently achieves superior performance across all benchmarks with better computational efficiency, while spatial-first also yields competitive results, confirming the robustness of our spatio-temporal pruning framework regardless of module ordering.

\section{Experiment}

\subsection{Experimental Details}

\subsubsection{Dataset and Metric} 
We evaluate ST-Prune on four mainstream autonomous driving 
benchmarks spanning complementary task categories, as summarized 
below:

\textbf{DriveLM}~\cite{drivelm} is a multi-view, single-frame benchmark built upon the nuScenes dataset, covering a broad spectrum of driving-relevant reasoning including cross-view risk target perception, risk target state prediction, and ego motion prediction. Evaluation is performed via four complementary metrics: \textit{Accuracy}, \textit{ChatGPT score}, \textit{Language score}, and \textit{Match score}. A comprehensive \textit{Average} score incorporating all four metrics is reported by the official online evaluation system.

\textbf{LingoQA}~\cite{lingoqa} is a single-view, multi-frame benchmark oriented toward ego-vehicle planning and driving reasoning. Unlike DriveLM, LingoQA focuses on temporal reasoning across consecutive frames, making it particularly suited for evaluating the temporal dimension of our framework. Model responses are evaluated using \textit{Lingo-Judge}, a learned text classifier specifically trained to assess the quality of driving-related natural language answers.

\textbf{NuInstruct}~\cite{nuinstruct} is a full multi-view, multi-frame benchmark encompassing a diverse range of high-level prediction and decision-making tasks. It is the most comprehensive benchmark in our evaluation suite, covering four distinct task types evaluated by dedicated metrics: \textit{Mean Absolute Error (MAE)} for regression tasks, \textit{Accuracy} for classification tasks, \textit{Mean Average Precision (mAP)} for object detection and grounding tasks, and \textit{BLEU} for captioning tasks. A unified \textit{Average} score is reported across all task types.

\textbf{OmniDrive}~\cite{omnidrive} is a full multi-view, multi-frame benchmark primarily focused on complex scene description and counterfactual reasoning tasks in diverse driving scenarios. Evaluation relies on rule-based natural language generation metrics.


\subsubsection{Foundation Model} 

The selection of an appropriate backbone is significantly constrained by the availability of open-source VLMs that natively support both multi-view and multi-frame inputs, which is a prerequisite for evaluating spatio-temporal token pruning. To the best of our knowledge, DriveMM\cite{drivemm} is currently the only publicly available model that satisfies this requirement while simultaneously demonstrating competitive performance across all four benchmarks adopted in this work. Its general-purpose structure—pairing a SigLIP~\cite{siglip} vision encoder with a Llama-3.1-8B~\cite{llama3} backbone—makes it an ideal platform to demonstrate the generalization and seamless transition of our pruning framework to future multi-modal models. Other recent VLM-based driving models\cite{vggdrive, nuinstruct, mpdrive} either lack open-source weights, support only single-view or single-frame inputs, or are architecturally specialized in ways that preclude fair comparison. Furthermore, both our internal evaluations and the results reported in the rexperiment of DriveMM~\cite{drivemm} indicate that direct supervised fine-tuning (SFT) of general-purpose models like Qwen\cite{qwen3-vl} on these datasets leads to a significant performance collapse. This makes DriveMM the most robust and technically rigorous baseline for assessing the impact of high-ratio token pruning in complex, multi-view driving scenarios.

\begin{table}[!t]
    \centering
    \footnotesize
    \renewcommand{\arraystretch}{1.3}
    \caption{Performance comparison under different token retention rates across four benchmarks. $^\dagger$ denotes the comprehensive scores provided by the official online evaluation system, which incorporates both the fundamental DriveLM metrics and detailed language generation scores (e.g., BLEU-1, ROUGE-L). $^*$ indicates (Accuracy + MAP + BLEU-MAE) / 4. }
    \label{tab:results}
    
    \vspace{0.2cm}
    \centerline{\textbf{(a) Remaining 25\% Tokens}}
    \vspace{0.1cm}
    \begin{tabularx}{\textwidth}{@{\extracolsep{\fill}}ll  c  cccc}
    \toprule
    \multirow{2}{*}{Dataset} & \multirow{2}{*}{Metric} & Full Tokens & \multicolumn{4}{c}{Pruning Methods} \\
    \cmidrule(l){3-3}\cmidrule(l){4-7}
    & & DriveMM (Vanilla) & \makecell{VisPruner \\ \footnotesize{\textit{ICCV 2025}}} & \makecell{DivPrune \\ \footnotesize{\textit{CVPR 2025}}} & \makecell{PACT \\ \footnotesize{\textit{CVPR 2025}}} & ST-Prune \\
    \hline
    \multirow{5}{*}{DriveLM} 
    & Accuracy$\uparrow$ & 81.23 & 75.84 & 76.72 & 78.26 & \textbf{81.10} \\
    & ChatGPT$\uparrow$ & 65.44 & \textbf{65.39} & 64.59 & 64.24 & 64.48 \\
    & Language$\uparrow$ & 51.33 & 51.37 & 50.03 & 49.84 & \textbf{52.41} \\
    & Match$\uparrow$ & 33.92 & 30.60 & 32.36 & 33.34 & \textbf{34.85} \\
    & Average$^\dagger\uparrow$ & 59.11 & 57.12 & 57.03 & 57.32 & \textbf{58.83} \\
    \hline
    LingoQA & LingoJudge$\uparrow$ & 70.80 & 65.40 & 62.80 & 64.00 & \textbf{68.20} \\ 
    \hline
    \multirow{5}{*}{NuInstruct} & MAE $\downarrow$ 
    & 3.50 & 3.68 & 3.78 & 10.82 & \textbf{3.49} \\
    & Accuracy $\uparrow$ 
    & 69.84 & 62.15 & 64.31 & 21.05 & \textbf{69.66} \\
    & mAP $\uparrow$ 
    & 22.32 & 1.85 & 0.00 & 0.00 & \textbf{24.34} \\
    & BLEU $\uparrow$ 
    & 24.72 & 30.12 & \textbf{33.17} & 22.50 & 25.74 \\
    & Average $\uparrow$ 
    & 28.35 & 22.61 & 23.43 & 8.17 & \textbf{29.06} \\
    \hline
    \multirow{4}{*}{OmniDrive}
    & BLEU$\uparrow$ & 40.78 & 26.94 & 39.59 & 37.39 & \textbf{40.82} \\
    & CIDEr$\uparrow$ & 112.65 & 66.52 & 108.76 & 98.51 & \textbf{113.24} \\
    & ROUGE$\uparrow$ & 39.79 & 32.63 & 39.05 & 37.55 & \textbf{39.76} \\
    & Average$^*\uparrow$ & 64.41 & 42.03 & 62.47 & 57.82 & \textbf{64.69} \\
    \bottomrule
    \end{tabularx}
    
    \vspace{0.6cm} 
    
    \centerline{\textbf{(b) Remaining 10\% Tokens}}
    \vspace{0.1cm}
    \begin{tabularx}{\textwidth}{@{\extracolsep{\fill}}ll c  cccc}
    \toprule
    \multirow{2}{*}{Dataset} & \multirow{2}{*}{Metric} & Full Tokens & \multicolumn{4}{c}{Pruning Methods} \\
    \cmidrule(l){3-3}\cmidrule(l){4-7}
    & & DriveMM (Vanilla) & \makecell{VisPruner \\ \footnotesize{\textit{ICCV 2025}}} & \makecell{DivPrune \\ \footnotesize{\textit{CVPR 2025}}} & \makecell{PACT \\ \footnotesize{\textit{CVPR 2025}}} & ST-Prune \\
    \hline
    \multirow{5}{*}{DriveLM} 
    & Accuracy$\uparrow$ & 81.23 & 74.15 & 73.22 & 76.31 & \textbf{77.60}\\
    & ChatGPT$\uparrow$ & 65.44 & 64.84 & \textbf{64.98} & 64.76 & 64.72\\
    & Language$\uparrow$ & 51.33 & 48.11 & 46.76 & 47.83 & \textbf{52.81} \\
    & Match$\uparrow$ & 33.92 & 30.02 & 30.91 & 32.94 & \textbf{33.57}\\
    & Average$^\dagger\uparrow$ & 59.11 & 55.93 & 55.85 & 56.81 & \textbf{57.87}\\
    \hline
    LingoQA & LingoJudge$\uparrow$ & 70.80 & 63.00 & 59.40 & 63.70 &  \textbf{65.20} \\ 
    \hline
    \multirow{5}{*}{NuInstruct} & MAE $\downarrow$ 
    & 3.50 & 3.76 & 3.89 & 11.24 & \textbf{3.52} \\
    & Accuracy $\uparrow$ 
    & 69.84 & 58.22 & 61.41 & 19.53 & \textbf{69.17} \\
    & mAP $\uparrow$ 
    & 22.32 & 1.05 & 0.00 & 0.00 & \textbf{19.74} \\
    & BLEU $\uparrow$ 
    & 24.72 & 31.84 & \textbf{34.09} & 21.47 & 25.74 \\
    & Average $\uparrow$ 
    & 28.35 & 21.84 & 22.90 & 7.44 & \textbf{27.78} \\
    \hline
    \multirow{4}{*}{OmniDrive}
    & BLEU$\uparrow$ & 40.78 & 19.43 & 39.45 & 28.52 & \textbf{40.78}\\
    & CIDEr$\uparrow$ & 112.65 & 48.84 & 107.21 & 69.36 & \textbf{112.79}\\
    & ROUGE$\uparrow$ & 39.79 & 28.37 & 38.78 & 32.67 & \textbf{39.81}\\
    & Average$^*\uparrow$ & 64.41 & 32.21 & 62.14 & 43.52 & \textbf{64.46}\\
    \bottomrule
    \end{tabularx}
\end{table}

\subsubsection{Comparative Baselines} 
We evaluate our proposed pruning framework against several representative training free pruning baselines comprising VisPruner\cite{vispruner}, DivPrune\cite{divprune}, and PACT\cite{pact}. These three methods encompass a diverse range of methodologies such as attention mechanisms, similarity based metrics, and clustering based strategies. Notably, the three baseline methods operate at distinct pipeline positions: VisPruner prunes within the vision encoder, DivPrune operates between the projector and LLM, and PACT executes pruning within the LLM layers, collectively covering the full spectrum of possible pruning positions in VLM pipelines. Additionally, we benchmark ST-Prune against Prune2Drive \cite{prune2drive}, the sole training-free competitor within AD domains. Due to its architectural limitation to multi-view processing, we evaluate this baseline solely on the DriveLM dataset. To ensure a rigorous and fair comparison, we assess the performance of our method across two widely adopted pruning regimes where 75\% and 90\% of the total visual tokens are removed on average. 

\subsubsection{Implementation Setup} 

All experiments are conducted on a single NVIDIA RTX PRO 6000 GPU 
with one batch size during inference. The balancing coefficients 
$\lambda_1$ and $\lambda_2$ are set to 0.6 and 0.8 respectively throughout all 
experiments.

\subsection{Main Results}

\noindent\textbf{Single-modality Benchmarks.} DriveLM and LingoQA independently validate the efficacy of RSP and MTP respectively, as each benchmark exercises a single spatio-temporal dimension. On DriveLM, ST-Prune demonstrates remarkable resilience to token compression, achieving the closest alignment to the Vanilla baseline with marginal degradation of only 0.28 and 1.24 points in Average score at 25\% and 10\% retention rates respectively, significantly narrower than competing methods. More notably, ST-Prune achieves 52.41 in Language score and 34.85 in Match score at 25\% retention, both surpassing the full-token Vanilla baseline. We attribute this counter-intuitive improvement to a noise regularization effect: by removing visually redundant tokens that may otherwise distract the language model, our cross-view pruning sharpens the model's focus on unique and informative spatial content, leading to more precise linguistic responses. On LingoQA, ST-Prune similarly leads all pruning baselines by a substantial margin, achieving 68.20 and 65.20 in Lingo-Judge score at 25\% and 10\% retention rates respectively, demonstrating that MTP effectively preserves the sequential reasoning cues essential for driving decision-making.

\noindent\textbf{Full Spatio-temporal Benchmarks.} NuInstruct and OmniDrive jointly validate the primary strength of ST-Prune as a unified spatio-temporal framework. On NuInstruct, ST-Prune retains an impressive 98.78\% of the original model performance at 25\% token retention, while competing methods fail to maintain scores above 80\%. The superiority is most pronounced in grounding tasks: while alternative methods yield near-zero mAP scores, ST-Prune achieves 24.34 mAP at 25\% retention and 19.74 mAP at 10\% retention, demonstrating its exceptional ability to preserve fine-grained spatial information that heuristic pruning fundamentally fails to maintain. This stark contrast suggests that without explicit spatio-temporal awareness, pruning methods inadvertently discard the precise localization cues required for object grounding. On OmniDrive, ST-Prune achieves near-identical performance to the full-token baseline across all language metrics at both retention rates, with Average scores of 64.69 and 64.47 compared to the Vanilla baseline of 64.41, further confirming that our framework maintains comprehensive scene understanding even under extreme compression.

\noindent\textbf{Comparison with AD-specific Pruning.} Table~\ref{tab:comparison_prune2drive} further validates ST-Prune against Prune2Drive~\cite{prune2drive}, the most closely related AD-specific baseline. To ensure a fair and rigorous comparison, we follow the same decimal point precision as reported in the original Prune2Drive paper. Although the absolute differences between the two methods appear modest at this precision level, ST-Prune consistently achieves higher Average scores under both retention regimes --- 58.8 vs. 58.3 at 25\% retention and 57.9 vs. 57.4 at 10\% retention --- and outperforms Prune2Drive across the majority of individual metrics in both settings. This consistent margin is particularly noteworthy given that Prune2Drive relies on an intensive Bayesian parameter search to identify optimal token allocations across camera views, whereas ST-Prune requires no offline search or additional training whatsoever. The results demonstrate that principled spatio-temporal constraints derived from domain geometry provide a more robust and practical foundation than data-driven search processes, achieving superior performance at zero additional deployment cost.

\noindent\textbf{Inference Efficiency.} Table~\ref{tab:efficiency_relative} reports throughput speedup relative to the DriveMM baseline across NuInstruct and OmniDrive, decomposed into prefill and decode phases. A notable observation concerns VisPruner, which achieves strong prefill speedups of 3.76$\times$ and 3.81$\times$ yet suffers severe decode slowdowns of 0.40$\times$ and 0.52$\times$ respectively. This stems from a fundamental pipeline difference: since VisPruner prunes tokens before the projector, it reduces the token count sufficiently early that the pipeline's native downsampling operation becomes unnecessary, resulting in a longer-than-expected token sequence entering the LLM decode phase. While this makes the comparison fair in terms of pipeline configuration, it reveals an inherent efficiency limitation of pre-projector pruning strategies that undermines their practical deployment value. Among methods that preserve the native pipeline, DivPrune achieves the highest raw end-to-end speedups of 3.18$\times$ and 2.36$\times$, yet this comes at a substantial performance cost on NuInstruct where it retains only 80.8\% of the original score. PACT similarly suffers severe performance degradation on both benchmarks despite moderate speedups, rendering it practically unsuitable for autonomous driving deployment. ST-Prune occupies a uniquely favorable position on the performance-efficiency Pareto frontier: it delivers consistent end-to-end speedups of over 2.5$\times$ on NuInstruct and approximately 2.0$\times$ on OmniDrive, while retaining 97.0\% and 100.1\% of the full-token baseline performance respectively — a combination unmatched by any competing method. The slight variation in speedup across datasets reflects differences in sequence length and task complexity, yet ST-Prune consistently maintains the best balance between computational efficiency and task performance across both settings.

\begin{table}[!t]
    \centering
    \footnotesize
    \renewcommand{\arraystretch}{1.3}
    \caption{Performance comparison with Prune2Drive baseline across different token retention rates. We follow the same decimal point precision as Prune2Drive.}
    \label{tab:comparison_prune2drive}
    \begin{tabularx}{\textwidth}{@{\extracolsep{\fill}}l ccccccc}
    \toprule
    Method & Accuracy  & ChatGPT  & BLEU-4  &Rouge  & CIDEr  & Match  & Average  \\
    \hline
    \rowcolor{gray!20} \multicolumn{8}{c}{Full Tokens}  \\
    Vanilla & 0.81 & 65.44 & 0.61 & 0.74 & 0.29 & 33.9 & 59.1 \\
    \rowcolor{gray!20} \multicolumn{8}{c}{Remain 25\% Tokens} \\
    Prune2Drive & 0.80 & \textbf{64.92} & 0.60 & 0.75 & 0.20 & 34.0 & 58.3 \\
    ST-Prune & \textbf{0.81} & 64.48 & \textbf{0.62} & \textbf{0.75} & \textbf{0.20} & \textbf{34.9} & \textbf{58.8} \\
    \rowcolor{gray!20} \multicolumn{8}{c}{Remain 10\% Tokens}  \\
    Prune2Drive & 0.78 & 64.52 & 0.56 & 0.74 & 0.16 & 33.4 & 57.4 \\
    ST-Prune & \textbf{0.78} & \textbf{64.72} & \textbf{0.61} & \textbf{0.74} & \textbf{0.23} & \textbf{33.6} & \textbf{57.9} \\
    \bottomrule
    \end{tabularx}
\end{table}

\begin{table}[!t]
    \centering
    \footnotesize
    \renewcommand{\arraystretch}{1.3}
    \caption{Efficiency comparison on NuInstruct and OmniDrive datasets. 
    Throughput metrics are presented as relative speedup factors compared 
    to the DriveMM baseline, where a value of 1.00 indicates baseline 
    performance. All methods are evaluated under the 10\% token retention 
    rate. Retention ratio is computed relative to the Vanilla baseline.}
    \label{tab:efficiency_relative}
    \begin{tabularx}{\textwidth}{@{\extracolsep{\fill}}ll ccc c}
    \toprule
    \multirow{2}{*}{Dataset} & \multirow{2}{*}{Method}  
    & \multicolumn{3}{c}{Throughput Speedup} 
    & \multirow{2}{*}{Avg. Score(Retention)} \\
    \cmidrule{3-5}
    & & Prefill & Decode & End-to-End & \\
    \midrule
    \multirow{5}{*}{NuInstruct} 
    & Vanilla   & 1.00$\times$ & 1.00$\times$ & 1.00$\times$ & 28.35 (100.0\%) \\
    & + VisPruner & 3.76$\times$ & 0.40$\times$ & 1.74$\times$ & 21.84 (77.0\%) \\
    & + DivPrune  & \textbf{4.24}$\times$ & \textbf{1.22}$\times$ & \textbf{3.18}$\times$ & 22.90 (80.8\%) \\
    & + PACT      & 2.56$\times$ & 1.05$\times$ & 2.36$\times$ & 8.17 (28.8\%) \\
    & + Ours      & 2.95$\times$ & 1.07$\times$ & 2.84$\times$ & \textbf{27.78 (97.99\%)} \\
    \midrule
    \multirow{5}{*}{OmniDrive} 
    & Vanilla   & 1.00$\times$ & 1.00$\times$ & 1.00$\times$ & 64.41 (100.0\%) \\
    & + VisPruner & 3.81$\times$ & 0.52$\times$ & 1.78$\times$ & 32.21 (50.0\%) \\
    & + DivPrune  & \textbf{4.81$\times$} & 1.01$\times$ & \textbf{2.36}$\times$ & 62.14 (96.5\%) \\
    & + PACT      & 2.08$\times$ & 0.99$\times$ & 1.61$\times$ & 43.52 (67.6\%) \\
    & + Ours      & 2.47$\times$ & \textbf{1.18$\times$} & 1.98$\times$ & \textbf{64.46 (100.1\%)} \\
    \bottomrule
    \end{tabularx}
\end{table}

\subsection{Ablation Study}

\noindent\textbf{Module Effectiveness.} As shown in Table~\ref{tab:ablation1}, both MTP and RSP independently deliver consistent improvements over vanilla max-min selection on the full spatio-temporal benchmarks. Notably, simultaneously adding temporal and spatial score achieves a total gain of +8.88\%, nearly matching the arithmetic sum of individual contributions (+9.23\%), which demonstrates that the two modules address largely complementary and non-overlapping sources of redundancy. 

\begin{table}[t]
    \centering
    \footnotesize
    \renewcommand{\arraystretch}{1.3}
    \caption{Ablation study on the contribution of each domain-specific scoring function. Average Gain is computed over NuInstruct and OmniDrive to ensure consistent comparison across all configurations, as these are the only two benchmarks where all module combinations are applicable. Dashes indicate benchmarks where the module combination is not applicable due to single-modality input constraints.}
    \label{tab:ablation1}
    \begin{tabularx}{\textwidth}{@{\extracolsep{\fill}}cc cccc c}
    \toprule
    \multicolumn{2}{c}{Ablation Modules} & 
    \multicolumn{2}{c}{Single-modality Benchmarks} & 
    \multicolumn{3}{c}{Spatio-temporal Benchmarks} \\
    \cmidrule(lr){1-2} \cmidrule(lr){3-4} \cmidrule(lr){5-7}
    
    Temporal  & Spatial  & DriveLM & LingoQA & NuInstruct & OmniDrive & Average Gain \\
    \hline

    \xmark & \xmark & 57.14 & 65.60 & 26.38 & 60.12 & - \\
    
    \cmark & \xmark & — & \textbf{68.20} & 27.45 & 62.45 & +3.97\% \\
    \xmark & \cmark & \textbf{58.83} & — & 27.84 & 63.12 & +5.26\% \\
    
    \cmark & \cmark & - & - & \textbf{29.06} & \textbf{64.69} & \textbf{+8.88\%} \\
    
    \bottomrule
    \end{tabularx}
\end{table}
\begin{table}[!t]
    \centering
    \footnotesize
    \renewcommand{\arraystretch}{1.3}
    \caption{Ablation study on the budget allocation under fixed total token retention rates of 25\% and 10\%.}
    \label{tab:intergrated}
   \begin{tabularx}{\textwidth}{@{\extracolsep{\fill}}ccc  cc}
    \toprule
    \multirow{2}{*}{Total Ratio}& \multicolumn{2}{c}{Detailed Retention Configuration} & \multicolumn{2}{c}{Average Score}\\
    \cmidrule(lr){2-3} \cmidrule(lr){4-5}
     & Temporal & Spatial
    & NuInstruct & OmniDrive \\
    \hline
    \multirow{3}{*}{Remain 25\% tokens} 
    & 100\% & 25\% & 27.59 & 64.19 \\
    & 50\%  & 50\% & \textbf{29.06} & \textbf{64.69} \\
    & 25\%  & 100\% & 28.09 & 64.10 \\
    \hline
    \multirow{5}{*}{Remain 10\% tokens} 
    & 100\% & 10\% & 24.98 & 63.96 \\
    & 50\%  & 20\% & 25.99 & \textbf{64.47} \\
    & 32\%  & 32\% & \textbf{27.78} & 64.46 \\
    & 20\%  & 50\% & 26.08 & 64.23 \\
    & 10\%  & 100\% & 26.65 & 64.03 \\
    \bottomrule
    \end{tabularx}
\end{table}

\begin{table}[!t]
    \centering    
    \footnotesize
    \renewcommand{\arraystretch}{1.3}
    \caption{Ablation study on pruning order under 25\% retention ratio.}
    \label{tab:ablation3}
    \begin{tabularx}{\textwidth}{@{\extracolsep{\fill}}c  cc  cc}
    \toprule
    \multirow{2}{*}{Order} & \multicolumn{2}{c}{NuInstruct} & \multicolumn{2}{c}{OmniDrive}\\ 
    \cmidrule(l){2-3} \cmidrule(l){4-5}
    & Avg. Score & E2E Latency(s/sample) & Avg. Score & E2E Latency(s/sample) \\
    \hline
    MTP $\rightarrow$ RSP & \textbf{29.06} & \textbf{1.71} & \textbf{64.69} & \textbf{2.83}\\
    RSP $\rightarrow$ MTP & 28.81 & 1.86 & 64.63 & 2.99\\
    \bottomrule
    \end{tabularx}
\end{table}
\begin{table}[t]
    \centering    
    \footnotesize
    \caption{Ablation study on the impact of pruning insertion position within the model pipeline.}
    \label{tab:position}
    \begin{tabularx}{\textwidth}{@{\extracolsep{\fill}}c  cccc  c}
    \toprule
    \multirow{2}{*}{Pruning Position} & 
    \multicolumn{4}{c}{Benchmarks} & 
    \multirow{2}{*}{Average Score} \\
    \cmidrule(lr){2-5}
     & DriveLM & LingoQA & NuInstruct & OmniDrive & \\ 
    \midrule
    ViT--Projector & 57.12& 66.20 & 22.80 & 61.56 & 51.92\\
    Projector--LLM & 56.39 & 61.60 & 23.58 & 62.46 & 51.00\\
     Inside LLM & \textbf{58.83} & \textbf{68.20} & \textbf{29.06} & \textbf{64.69} & \textbf{54.67} \\
    \bottomrule
    \end{tabularx}
\end{table}

\begin{figure}[t]
  \centering
  \includegraphics[width=0.95\linewidth]{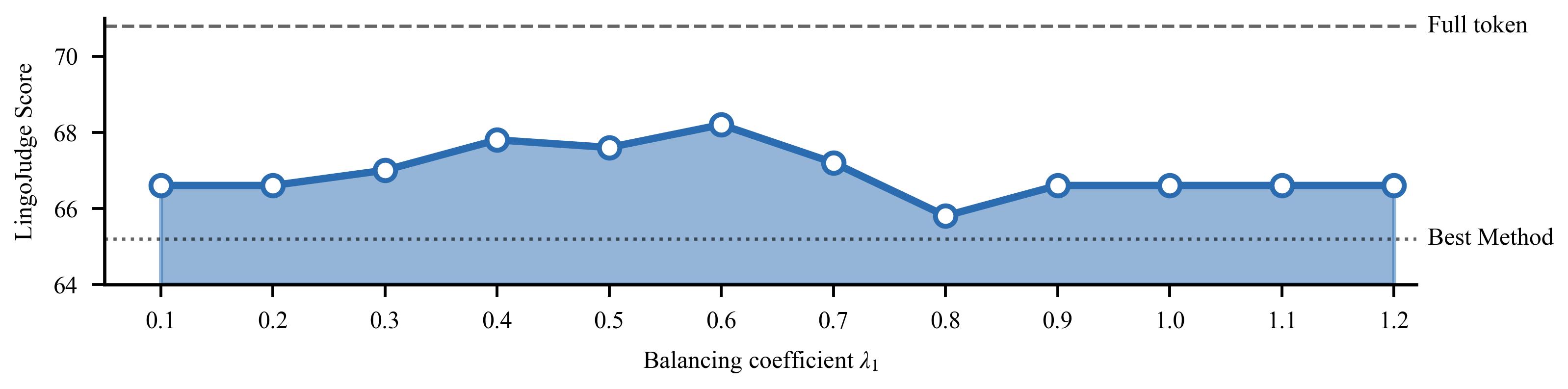} \\
  
  
  \includegraphics[width=0.95\linewidth]{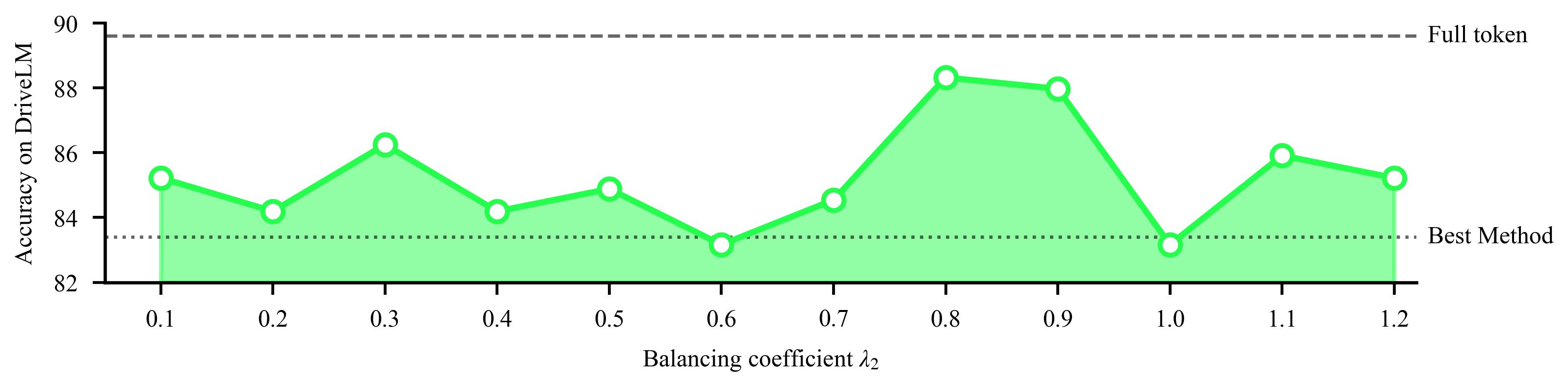}
  
  \caption{Sensitivity analysis of the balancing coefficients $\lambda_1$ (top) and $\lambda_2$ (bottom) for MTP on LingoQA and RSP modules on DriveLM, respectively}
  \label{fig:lambda}
\end{figure}

\noindent\textbf{Budget Allocation.} Table~\ref{tab:intergrated} investigates the budget allocation between MTP and RSP under fixed total token budgets of 25\% and 10\%. A balanced allocation consistently yields the strongest overall performance: at 25\% budget, the 50\%+50\% split achieves the highest average scores on both benchmarks, while at 10\% budget, the symmetric 32\%+32\% split delivers the best NuInstruct average of 27.78, confirming that neither temporal nor spatial pruning should dominate. Notably, over-relying on a single dimension consistently degrades performance, particularly on grounding tasks where mAP collapses significantly under extreme single-dimension compression. Nevertheless, even under these suboptimal single-dimension configurations, ST-Prune consistently outperforms all general-purpose pruning baselines reported in Table~\ref{tab:results}, demonstrating that the domain-specific spatio-temporal awareness embedded in each module alone is sufficient to surpass methods that lack such structural priors entirely.

\noindent\textbf{Pruning Order.} We empirically compare both orders in Table~\ref{tab:ablation3}: the temporal-first order (MTP$\rightarrow$RSP) achieves marginally superior performance on both benchmarks while delivering a consistent latency reduction of 0.15--0.16 seconds per sample, confirming our theoretical complexity analysis that temporal-first incurs lower overhead by reducing the token count before spatial similarity precomputation. Notably, the spatial-first order also yields competitive results, demonstrating the robustness of ST-Prune's spatio-temporal pruning framework regardless of module ordering.

\noindent\textbf{Pruning Postion. }Motivated by the observation that existing pruning methods operate at fundamentally different pipeline positions, we conduct a controlled position study by applying ST-Prune at three candidate locations: between the vision encoder and projector, between the projector and LLM, and within the LLM phase. As shown in Table~\ref{tab:position}, pruning inside the LLM consistently outperforms both earlier positions by a substantial margin, confirming that LLM-phase token representations carry richer semantic attention that enables more meaningful diversity-based selection. This result also provides a unified explanation for the performance gap between ST-Prune and general-purpose baselines: beyond methodological differences, positional alignment with the LLM phase is itself a critical factor in preserving task performance. 

\noindent\textbf{Parameters Analysis.} Figure~\ref{fig:lambda} presents the sensitivity analysis of the balancing coefficients $\lambda_1$ and $\lambda_2$ for MTP and RSP respectively. To directly isolate the effect of each module, we validate on single-dimension benchmarks: $\lambda_1$ is evaluated on LingoQA using the standard Lingo-Judge metric, while $\lambda_2$ is evaluated on a dedicated offline subset constructed from DriveLM 1.0 using Accuracy as the representative metric due to the online evaluation frequency limitation.  Full spatio-temporal benchmarks such  as NuInstruct are deliberately excluded from this analysis, as their performance reflects the joint effect of both $\lambda_1$ and $\lambda_2$ simultaneously, which would confound the individual sensitivity measurements. Two consistent observations emerge across both figures. First, performance degrades when the balancing coefficient is either too small or too large: an insufficient value renders the domain-specific scoring function ineffective, causing the selection to degenerate toward vanilla max-min diversity, while an excessive value allows the scoring function to overwhelm feature diversity, suppressing the complementary information that diversity selection is designed to capture. For MTP, the optimal stable range is $\lambda_1 \in [0.4, 0.6]$, while RSP requires a stronger penalty with an optimal range of $\lambda_2 \in [0.8, 0.9]$, reflecting the fact that bilateral cross-view similarity constitutes a weaker signal than temporal variance and therefore demands greater emphasis to effectively influence the selection. Despite these performance variations across the parameter range, ST-Prune consistently outperforms the best competing pruning method across virtually all evaluated values of both $\lambda_1$ and $\lambda_2$, as indicated by the dotted reference line in each figure. This demonstrates that ST-Prune is not critically sensitive to the precise choice of these hyperparameters, since the domain-specific spatio-temporal structural priors embedded in MTP and RSP provide a robust performance advantage over structure-agnostic baselines even under suboptimal parameter settings.

\begin{figure}[!t]
\centering
\includegraphics
[width=0.95\linewidth]
{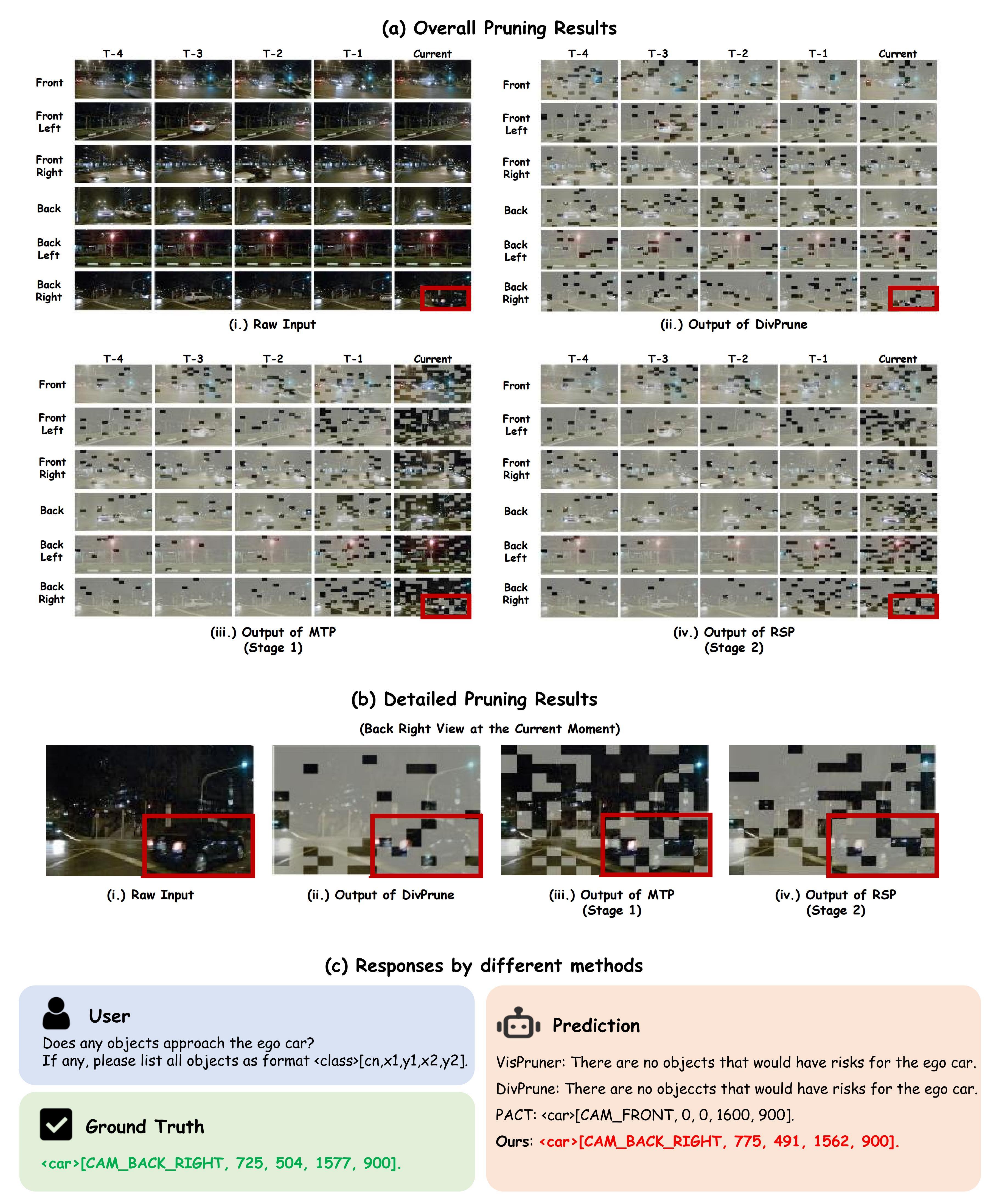}
\caption{Qualitative visualization of ST-Prune on a challenging 
grounding scenario from NuInstruct. \textbf{(a)} Overall pruning 
results across all six camera views and five consecutive frames 
(T-4 to Current), comparing raw input, DivPrune, and the two 
stages of ST-Prune (MTP and RSP). The red bounding box highlights 
the target vehicle in the Back Right camera at the current 
timestep. \textbf{(b)} Detailed token retention patterns zoomed 
into the Back Right view at the current moment \textbf{(c)} Responses from different pruning 
methods to the grounding query, where only ST-Prune correctly 
identifies the target camera and produces accurate bounding box 
coordinates.}
\label{fig:visualization}
\end{figure}

\subsection{Qualitative Analysis}

Figure~\ref{fig:visualization} presents a qualitative comparison on a challenging grounding scenario from NuInstruct, where the target object --- a vehicle approaching the ego car --- appears exclusively in the Back Right camera at the current timestep among a total of 30 spatio-temporal token maps spanning six camera views and five consecutive frames, posing a demanding localization challenge due to its small size and highly peripheral position.

Panel (a) reveals a striking visual contrast between traditional pruning methods represented by DivPrune and ST-Prune in their token allocation strategies across all six camera views and five consecutive frames. As shown in (ii.), DivPrune retains a roughly uniform proportion of tokens across all frames and viewpoints, treating foreground objects and static background regions indiscriminately --- historical frames T-4 through T-1 receive comparable token budgets to the current frame, and repetitive background scenery competes equally with safety-critical foreground content. In stark contrast, ST-Prune adopts a principled coarse-to-fine strategy. As shown in (iii.), MTP first aggressively filters the tremendous repeated background details accumulated across historical frames, concentrating the token budget toward the current frame where the target vehicle resides. However, since temporal recency retains all current-frame content indiscriminately, some residual background features inevitably persist in the current view. RSP then effectively addresses this by penalizing tokens that appear redundantly across neighboring camera views, systematically suppressing the cross-view background overlap while preserving the spatially unique content around the target region, as shown in (iv.). Through this spatio-temporal pipeline, ST-Prune progressively concentrates the token budget onto the most informative and unique content, achieving a structured and coherent representation that DivPrune's flat selection fundamentally cannot replicate.

Panel (b) zooms into the Back Right view at the current timestep, making the progressive refinement of each stage clearly visible at the token level. Notably, both DivPrune as shown in (ii.) and ST-Prune as shown in (iii.) retain tokens in the vicinity of the target vehicle within the red bounding box, suggesting that token coverage alone is not the determining factor for task success. Nevertheless, it is evident that the token budget for this key scene related to concrete task in panel(c) by DivPrune is less lower than Our methods. MTP (iii) already achieves noticeably richer coverage of the target region through temporal recency bias, while RSP (iv) further sharpens this by concentrating the remaining budget around the spatially unique content within the red bounding box, producing the most coherent local context around the approaching vehicle. 

The practical consequence of these token-level differences is directly reflected in the grounding responses shown in panel (c). Although DivPrune retains a comparable number of tokens near the target vehicle, the useful information is distracted and diluted by the surrounding unstructured background tokens, preventing the language model from correctly interpreting the scene --- both VisPruner and DivPrune respond that there are no risks for the ego car, a dangerous failure in a safety-critical scenario. PACT, despite producing a bounding box, assigns it to the wrong camera \texttt{CAM\_FRONT} with degenerate coordinates \texttt{[0, 0, 1600, 900]}, indicating a complete collapse of spatial localization. ST-Prune alone correctly identifies \texttt{CAM\_BACK\_RIGHT} and produces the closest bounding box coordinates \texttt{[775, 491, 1562, 900]} to the ground truth \texttt{[725, 504, 1577, 900]}, demonstrating that coherent spatio-temporal context, rather than mere token retention, is the critical enabler of precise spatial grounding under extreme token compression.

\section{Conclusion}
In this paper, we present ST-Prune, a training-free, plug-and-play spatio-temporal token pruning framework for Vision-Language Models in autonomous driving. Motivated by the fundamental inability of existing token pruning methods to exploit the structured redundancy patterns inherent in multi-view, multi-frame driving scenarios, ST-Prune extends the principled max-min diversity selection framework with two complementary domain-specific modules. Motion-aware Temporal Pruning (MTP) incorporates motion volatility and temporal recency constraints to systematically prioritize dynamic trajectories and recent-frame content, while Ring-view Spatial Pruning (RSP) exploits the ring-view arrangement to eliminate duplicate cross-view projections and residual static background that temporal pruning alone cannot resolve. Operating in a coarse-to-fine pipeline, the two modules address spatio-temporal redundancy synergistically, each delivering consistent improvements as a standalone component while achieving superior performance in combination. Extensive experiments across four mainstream autonomous driving benchmarks demonstrate that ST-Prune consistently outperforms all existing pruning methods across perception, prediction, and planning tasks, retaining near-lossless performance under aggressive token compression while delivering end-to-end speedups comparable to existing pruning approaches. Furthermore, through extensive ablation studies and rigorous empirical analysis, we have substantiated the efficacy, robustness, and theoretical soundness of MTP and RSP. Our results underscore their critical roles both as standalone modules and as synergistic components within the unified ST-Prune framework, confirming its viability for high-efficiency autonomous driving.

Despite its strong empirical performance, ST-Prune has two primary limitations that point to promising directions for future work. First, while NuInstruct incorporates trajectory prediction tasks that partially cover VLA scenarios, the generalizability of ST-Prune to dedicated VLA datasets and closed-loop action generation benchmarks remains unexplored. Extending ST-Prune to such settings, where precise temporal ordering of action sequences is critical, represents an important next step toward deploying efficient VLMs in real-world autonomous driving stacks. Second, the scope of our evaluation is constrained by the limited availability of publicly accessible vision-language driving models that natively support multi-view and multi-frame inputs. Although we conducted additional experiments on a supervised fine-tuned model based on Qwen, the significantly lower baseline performance on our benchmarks compared to DriveMM makes fair comparison difficult, as performance gaps between pruning methods may reflect differences in base model capability rather than pruning effectiveness. As more capable and publicly available multi-view, multi-frame VLMs emerge, broader evaluation across diverse model architectures will be an important direction for future work.

\bibliographystyle{unsrt}  
\bibliography{ref}  


\end{document}